\documentclass[lettersize,journal]{IEEEtran}
\usepackage{amsmath,amsfonts}
\usepackage{algorithmic}
\usepackage{algorithm}
\usepackage{array}
\usepackage[caption=false,font=normalsize,labelfont=sf,textfont=sf]{subfig}
\usepackage{textcomp}
\usepackage{stfloats}
\usepackage{url}
\usepackage{verbatim}
\usepackage{multirow}
\usepackage{graphicx}
\usepackage{makecell}
\usepackage{cite}
\usepackage{booktabs}
\usepackage{xcolor}
\definecolor{citeblue}{HTML}{3279BD}
\makeatletter
\newcommand{\cvprtablecaptionfull}[1]{%
  \refstepcounter{table}%
  \noindent\parbox{\textwidth}{\small\textbf{Table \thetable.} #1}\par
}

\newcommand{\cvprtablecaptionhalf}[1]{%
  \refstepcounter{table}%
  \noindent\parbox{\columnwidth}{\small\textbf{Table \thetable.} #1}\par
}
\makeatother

\newcommand{\rotheader}[1]{\rotatebox[origin=c]{90}{\makecell[c]{#1}}}
\usepackage[colorlinks=true,
            linkcolor=blue,
            citecolor=citeblue,
            urlcolor=magenta]{hyperref}
\usepackage{cleveref}            
\begin{document}
\bstctlcite{BSTcontrol}
\title{KGEdit: Ambiguity-Aware Knowledge Graphs for Training-Free Precise Video Generation and Editing}

\author{{
Mingshu~Cai,
Miao~Zhang,
Chenghe~Yang,
Yixuan~Li,
Osamu~Yoshie, and~Yuya~Ieiri}
\thanks{Mingshu~Cai, Osamu~Yoshie, and Yuya~Ieiri are with Waseda University, Japan (e-mail: mignshucai@fuji.waseda.jp; yoshie@waseda.jp; ieyuharu@aoni.waseda.jp).}
\thanks{Miao~Zhang is with the College of Computer Engineering, Jimei University, Xiamen, Fujian, China (e-mail: zhangmiao@jmu.edu.cn).}
\thanks{Chenghe~Yang is with the Department of Language Science and Technology, The Hong Kong Polytechnic University, Hong Kong, China (e-mail: chenghe.yang@connect.polyu.hk).}
\thanks{Yixuan~Li is with the College of Computing and Data Science, Nanyang Technological University, Singapore (e-mail: yixuan.li@ntu.edu.sg).}
\thanks{Osamu~Yoshie is the corresponding author.}
}

\markboth{Journal of \LaTeX\ Class Files,~Vol.~14, No.~8, August~2021}%
{Shell \MakeLowercase{\textit{et al.}}: A Sample Article Using IEEEtran.cls for IEEE Journals}


\IEEEaftertitletext{%
\vspace{-1.8\baselineskip}
\begin{center}
\includegraphics[width=\textwidth]{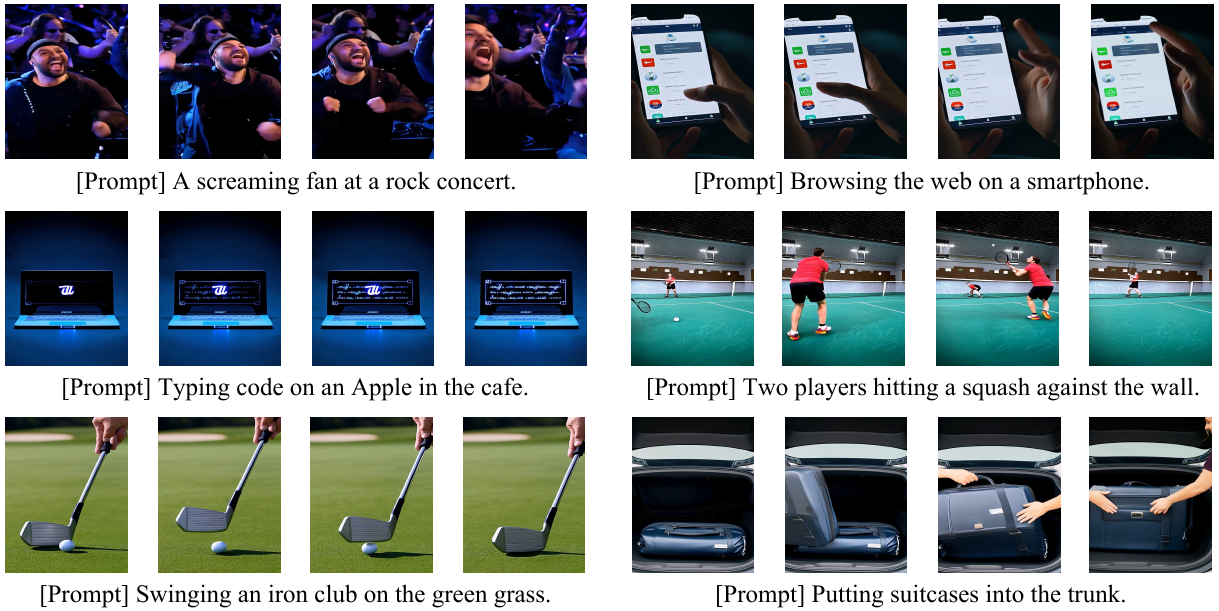}
\end{center}
\vspace{-0.5em}
\refstepcounter{figure}
\noindent\begin{minipage}{\textwidth}
\footnotesize
\textbf{Fig.~\thefigure.}
Our model is a unified video generation and editing framework that, through semantic disambiguation and precise attribute injection, can produce highly fine-grained and well-aligned video results within only a few rounds of user interaction.
\label{fig:teaser}
\end{minipage}
\vspace{0.2\baselineskip}
}

\maketitle

\begin{abstract}
In recent years, training-free video generation has progressed remarkably. However, when handling complex textual instructions, existing methods still suffer from semantic ambiguity, incorrect concept binding, and cross-frame inconsistency. To address these issues, we propose KGEdit, a structured semantic control framework for text-to-video (T2V) diffusion models. Specifically, we first construct an ambiguity-aware knowledge graph (AAKG) to disentangle and disambiguate the input prompt, converting it into four types of structured semantics: identity, relation, attribute, and negative constraints. We then design a structured semantic injection module (SSIM) to inject these semantic signals into key layers of the diffusion Transformer, enabling fine-grained semantic control. In addition, we introduce a temporal-aware semantic control (TASC) module that dynamically schedules semantic objectives according to the stage-wise characteristics of the denoising process, further improving semantic alignment and temporal consistency. Experiments show that KGEdit outperforms existing methods in editing precision and temporal stability, while offering higher efficiency and controllability in text-driven interaction scenarios. Our codebase is available at \url{https://github.com/CIMASA/KGEdit}.
\end{abstract}

\begin{IEEEkeywords}
Video generation and editing, knowledge graph, training-free framework, diffusion Transformer
\end{IEEEkeywords}

\section{Introduction}
\IEEEPARstart{W}{ith} the rapid development of diffusion models and large-scale video generative models, video generation and editing have achieved remarkable progress in recent years \cite{ho2020ddpm,ddim,jin2024flowmatching,peebles2023dit,makeav,blattmann2023align,cogvideox,hunyuanvideo,LTXVideo,chen2025goku}. The research paradigm has gradually evolved from early text-to-video generation toward more complex and flexible settings, including reference-image-guided generation, reference-video-guided generation, region-level control, and multi-condition compositional editing \cite{tuneav,vgenxl,videocomposer,chen2023controlavideo,wang2024motionctrl,liu2024videop2p,qi2023fatezero,geyer2024tokenflow,cong2024flatten,fluencyve,gao2026contro,nova}. This trend has further promoted the development of unified video creation frameworks, making it possible for a single model to support diverse tasks \cite{vace,wei2026vogue}, while also broadening the creative freedom and application potential of video content creation in domains such as advertising, animation, and gaming \cite{texttoedit,motioncanvas,gamefactory,anisora}.

Although existing methods have expanded task coverage and improved generation quality, and recent studies have evolved from task-specific pipelines toward unified video creation frameworks \cite{vace,wei2026vogue}, text remains the fundamental, natural, and accessible input modality for video generation and editing. Stronger control signals, such as reference videos, local masks, or sparse guidance, are introduced only when needed \cite{vace,wei2026vogue,gao2026contro,nova}. However, natural language is inherently ambiguous, underspecified, and coarse-grained, especially when requests involve identity, appearance, motion, local-region modifications, or joint control of multiple attributes. Existing methods rely on prompts, cross-attention manipulation, or inversion-based guidance \cite{liu2024videop2p,qi2023fatezero,wang2023zero,ku2024anyvv,he2026geadapter}. Yet these strategies often struggle to explicitly capture the key attributes and semantic relations governing the target video. Under complex instructions, models remain prone to semantic ambiguity, concept misbinding, and temporal instability. Achieving satisfactory results often requires repeated prompt refinement or additional control conditions, increasing both application complexity and inference cost \cite{gao2026contro,nova,zhang2024controlvideo}.

Despite improving cross-task and cross-modal compatibility \cite{vace,wei2026vogue}, unified video creation frameworks still struggle with precise semantic control and temporal stability under complex textual instructions. First, text conditions are usually encoded as holistic semantics rather than explicit structured components such as identity, relations, attributes, and negative constraints. But such structure helps only if it is injected effectively during denoising: naively enforcing semantic supervision throughout diffusion is unreliable, especially at early noisy timesteps, leading to weak attribute injection, concept confusion, and semantic drift \cite{liu2024videop2p,qi2023fatezero,ku2024anyvv,he2026geadapter}. Second, video generation requires not only plausible frame-level content but also stable propagation of key semantics over time; once control becomes unstable, errors readily accumulate as cross-frame inconsistency, flickering, and drift \cite{geyer2024tokenflow,cong2024flatten,fluencyve}. Moreover, different semantic components matter at different denoising stages, so uniform guidance cannot simultaneously support early structural grounding, mid-stage relation modeling, and late detail refinement. Therefore, achieving more accurate structured semantic understanding and more stable temporal control during denoising, without additional training, remains a key challenge for practical video generation and editing systems.

Motivated by these observations, we propose a training-free structured semantic control framework for text-to-video diffusion models. The core idea is that, rather than continually increasing input modalities or relying on repeated prompt refinement, it is more effective to improve the interpretability and controllability of the textual condition itself. Specifically, we first leverage an ambiguity-aware knowledge graph to disambiguate and structurally organize the key semantics in user instructions, and further decompose them into four semantic groups, namely identity, relation, attribute, and negative constraints, thereby alleviating ambiguity in natural language and highlighting the core semantics that truly determine the target video content \cite{bollacker2008freebase,vrandevcic2012wikidata,bordes2013transe}. On this basis, we design a structured semantic injection module to inject these structured semantic signals into key layers of the diffusion Transformer in a training-free manner, enabling finer-grained semantic control without additional model adaptation. Furthermore, we introduce a temporal-aware semantic control module, which dynamically schedules different semantic objectives according to the stage-wise characteristics of the denoising process, thereby improving both semantic alignment and temporal consistency. Through this design, our framework is able to produce more accurate, stable, and coherent video results under complex textual instructions. Representative examples are shown in Fig.~\ref{fig:teaser}. Overall, the main contributions of this work are as follows:

\begin{itemize}
    \item We propose \textbf{KGEdit}, a unified training-free framework for structured semantic control in video generation and editing. By improving the interpretability and controllability of textual conditions, KGEdit enables more accurate and temporally stable video generation under complex textual instructions, without requiring additional model training.

    \item We propose an \textbf{Ambiguity-Aware Knowledge Graph (AAKG)} and a \textbf{Structured Semantic Injection (SSIM)} mechanism to achieve structured semantic modeling and fine-grained control for complex textual instructions. These designs improve semantic understanding, semantic alignment, and generation accuracy.

    \item We propose a \textbf{Temporal-Aware Semantic Control (TASC)} module. Through dynamic semantic scheduling over the denoising process, TASC improves the temporal stability of key semantics during video generation.

    \item Extensive experiments verify the effectiveness of the proposed method. The results show that KGEdit outperforms existing methods in overall performance and achieves state-of-the-art results on multiple metrics of the VBench benchmark~\cite{huang2023vbench}.
\end{itemize}

\section{Related Works}
\subsection{Video Generation and Editing.}
Video generation and editing have advanced rapidly in recent years~\cite{cogvideox,gao2026contro,hunyuanvideo,LTXVideo,nova,vace,vgenxl}, enabling high-quality and controllable content creation for advertising, film, gaming, and animation~\cite{texttoedit,motioncanvas,gamefactory,anisora}. Early studies mainly adapted pretrained T2I diffusion models to the video domain to establish basic T2V generation capabilities~\cite{blattmann2023align,makeav}. Tune-A-Video~\cite{tuneav} was among the first to demonstrate one-shot text-driven video generation by extending pretrained T2I models with tailored spatio-temporal attention and DDIM inversion. Subsequent research shifted from basic generation toward more controllable video synthesis. Representative methods such as VideoComposer~\cite{videocomposer} and Control-A-Video~\cite{chen2023controlavideo} introduced diverse control signals, including text, sketch, depth, edge, and motion cues, while MotionCtrl~\cite{wang2024motionctrl} further disentangled camera motion and object motion for more precise manipulation. Meanwhile, video generation backbones have evolved toward stronger architectures and larger-scale foundation models. Latte~\cite{ma2025ditv} was among the early attempts to introduce latent diffusion Transformers into video generation, and more recent large-scale models such as Movie Gen, Wan, and Open-Sora 2.0~\cite{moviegen,wan2025wan,open_sora} have further improved the quality, scalability, and efficiency of video synthesis. In parallel, flow-based paradigms have shown increasing promise: Pyramidal Flow Matching~\cite{jin2024flowmatching} improves efficiency through a unified pyramidal flow formulation, while Goku~\cite{chen2025goku} demonstrates the potential of rectified-flow Transformers for joint image and video generation.

Alongside video generation, video editing methods have also progressed substantially. Video-P2P~\cite{liu2024videop2p} and FateZero~\cite{qi2023fatezero} extend cross-attention-based and inversion-based zero-shot editing from image editing to videos, enabling textual modifications of real videos while largely preserving their original layout and motion. Compared with image editing, video editing additionally requires temporal consistency across frames. To address this challenge, TokenFlow, FLATTEN, GE-Adapter and FluencyVE~\cite{geyer2024tokenflow,cong2024flatten,he2026geadapter,fluencyve} mitigate flickering and semantic drifting through diffusion feature correspondence, optical-flow-guided attention, lightweight adaptor-based temporal-spatial-semantic consistency modeling and temporal-aware Mamba-based sequence modeling, respectively. Building on these advances, VACE~\cite{vace} unifies video generation and editing within a shared framework, reflecting a broader shift from task-specific models to unified video creation systems.

\subsection{Knowledge Graphs.}
Knowledge graphs (KGs) were originally introduced to organize large-scale structured knowledge as entities and relations. Early systems such as Freebase~\cite{bollacker2008freebase} and Wikidata~\cite{vrandevcic2012wikidata} established machine-readable and extensible knowledge bases, enabling structured retrieval and reasoning in open-world settings. Later, KG research shifted toward representation learning, where TransE~\cite{bordes2013transe} became a seminal method for modeling multi-relational data with translational embeddings.

Recently, KGs have played an increasingly important role in visual and multimodal research, extending from external knowledge augmentation to structured visual representation and reasoning enhancement. VQA-GNN~\cite{wang2023vqa} supports collaborative reasoning over scene graphs, concept graphs, and question context, while From Pixels to Graphs~\cite{from_pix_g} advances open-vocabulary scene graph generation with vision-language models. VaLiK~\cite{Liu_2025_VaLiK} and ReasonVQA~\cite{tran2025reasonvqa} further promote KG-based visual understanding through multimodal KG construction and structured reasoning benchmarks. As video tasks become more multimodal and interactive, resolving ambiguity in user instructions motivates our use of KGs.

\subsection{Training-Free Visual Generative Model.}
Training-free visual generation and editing first emerged in the image domain, where pretrained diffusion models are manipulated at inference time to enable controllable synthesis and editing without additional fine-tuning. Existing image-based methods mainly improve editing flexibility, generation fidelity, and zero-shot customization through attention control, feature injection, and test-time adaptation strategies~\cite{hertz2023prompttoprompt,tumanyan2023plug,cao2023masactrl,feng2026personalize}. Among them, Stable Flow~\cite{avrahami2025stable} improves editing stability by identifying critical layers in DiTs, KV-Edit~\cite{zhu2025kv} enhances background consistency through KV-cache reuse, and FreeCus~\cite{zhang2025freecus} further demonstrates the potential of subject-driven customization without additional training.

This paradigm has also been rapidly extended to the video domain. Recent methods show that temporally coherent video generation and editing can be achieved in zero-shot or training-free manners by reusing pretrained backbones with motion guidance, image conditioning, inversion, and image-to-video propagation~\cite{zhang2024controlvideo,ni2024ti2v,wang2023zero,ku2024anyvv}. Representative examples include Text2Video-Zero~\cite{khachatryan2023t2video}, DirecT2V~\cite{hong2023large}, Free-Bloom~\cite{huang2023free}, and EIDT-V~\cite{jagpal2025eidt} for training-free video generation, as well as AnyV2V~\cite{ku2024anyvv} for video editing without additional tuning. As the cost of model adaptation continues to decrease, practical bottlenecks increasingly shift to the application side, especially in multimodal settings where ambiguous user inputs can limit controllability and faithfulness, motivating our focus on ambiguity-aware input understanding.

\section{Method}

\begin{figure*}[t]
    \centering
    \includegraphics[width=0.96\linewidth]{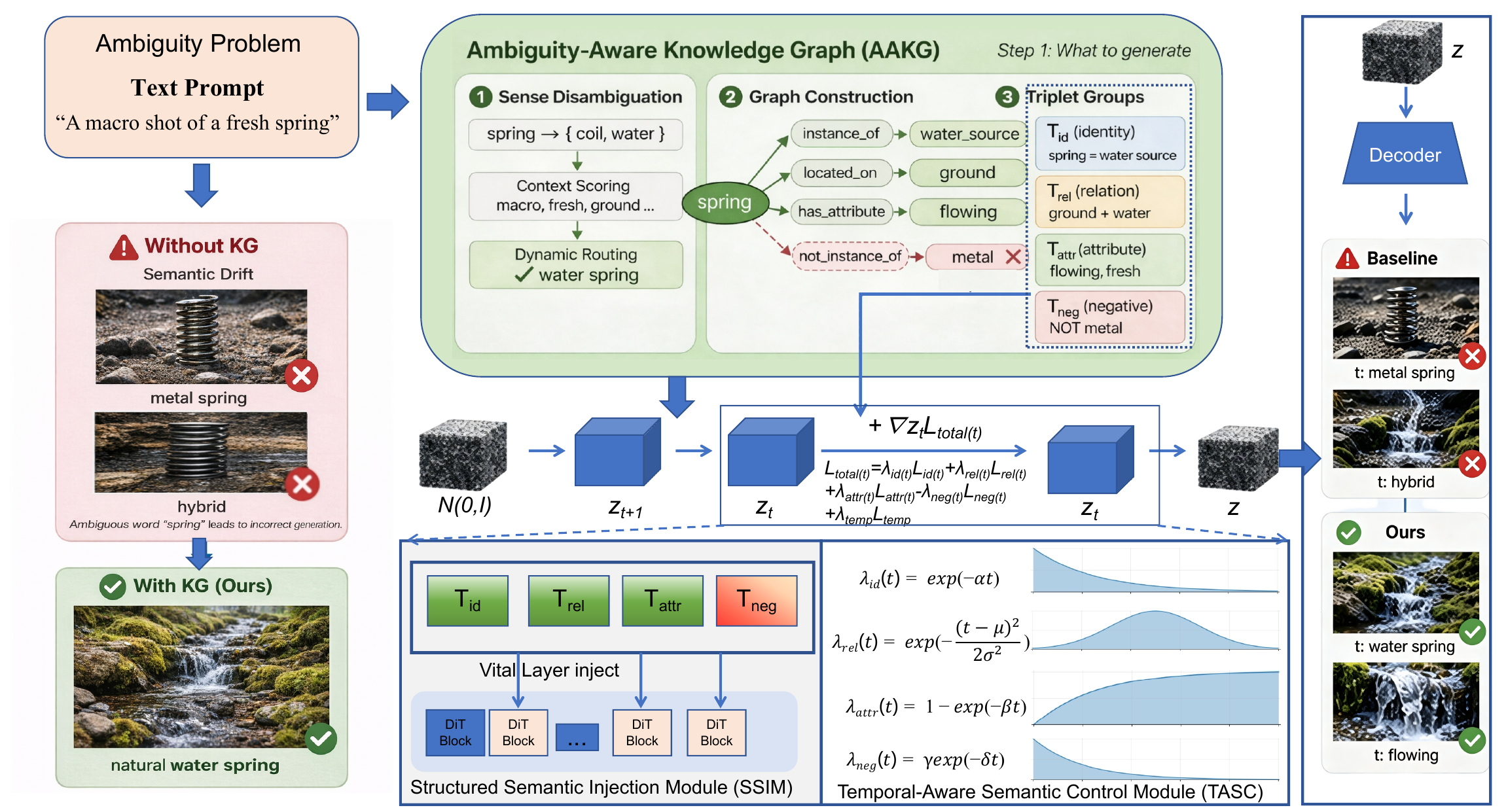}
    \vspace{-0.2cm}
    \caption{
    Overview of our ambiguity-aware diffusion framework.
    Given an ambiguous text prompt, we first perform sense disambiguation and construct an \emph{Ambiguity-Aware Knowledge Graph} (AAKG), which organizes prompt semantics into four triplet groups: identity, relation, attribute, and negative constraints.
    Based on these semantic groups, we design a \emph{Structured Semantic Injection Module} (SSIM) to inject structured guidance into selected DiT blocks, and a \emph{Temporal-Aware Semantic Control Module} (TASC) to adaptively schedule different semantic objectives across denoising timesteps.
    During sampling, the overall guidance objective is composed of identity, relation, attribute, negative, and temporal terms, whose gradients are used to steer the latent reverse process toward semantically consistent generation.
    Compared with vanilla generation under ambiguous prompts, our framework resolves semantic drift and produces outputs that better match the intended meaning.
    }
    \vspace{-0.4cm}
    \label{fig:framework}
\end{figure*}

\textbf{Overview.}
We propose a training-free framework for ambiguity-aware and structured semantic control in diffusion models. An overview of the proposed framework is shown in Fig.~\ref{fig:framework}.

Given a text prompt $\mathbf{y}$, conventional text-conditioned diffusion methods often rely on a single global embedding, 
which struggles to resolve semantic ambiguity and fails to enforce structured relationships, leading to semantic drift or incorrect concept binding. To address this issue, we first construct an Ambiguity-Aware Knowledge Graph (AAKG) to resolve semantic ambiguity and organize the input prompt into structured semantics. Based on this representation, we introduce a Structured Semantic Injection Module (SSIM) to incorporate structured semantic signals into key layers of the diffusion model for effective fine-grained control. Furthermore, to account for the stage-dependent roles of different semantic components during denoising, we design a Temporal-Aware Semantic Control (TASC) module that dynamically schedules semantic guidance over diffusion timesteps.

Overall, our framework unifies semantic disambiguation, structured representation, and temporal control into a single pipeline, 
enabling precise, consistent, and interpretable generation. 

\vspace{0.2cm}

\subsection{Preliminaries}

A diffusion model defines a gradual noising process together with a learned denoising process \cite{ho2020ddpm,sd}. Starting from a clean sample $\mathbf{z}_0 \sim p(\mathbf{z}_0)$, the forward process perturbs the data over $t \in \{1,\dots,T\}$ according to
\begin{equation}
	\mathbf{z}_t = \sqrt{\bar{\alpha}_t}\mathbf{z}_0 + \sigma_t \boldsymbol{\epsilon}_t,
\end{equation}
where $\bar{\alpha}_t = \prod_{i=1}^{t}(1-\beta_i)$, $\sigma_t = \sqrt{1-\bar{\alpha}_t}$, and $\boldsymbol{\epsilon}_t \sim \mathcal{N}(0,\mathbf{I})$. Here, $\beta_t$ denotes the variance schedule controlling the noise intensity at each timestep. Under this construction, $\mathbf{z}_t$ follows a Gaussian distribution centered at $\sqrt{\bar{\alpha}_t}\mathbf{z}_0$.

The reverse process aims to recover clean data from noisy samples. In practice, a neural network $\boldsymbol{\epsilon}_{\boldsymbol{\theta}}$ is trained to estimate the injected noise:
\begin{equation}
\label{eq:ddpm_loss}
\min_{\boldsymbol{\theta}} 
\mathbb{E}_{\mathbf{z}_t,\boldsymbol{\epsilon}_t,t}
\left[
\left\|
\boldsymbol{\epsilon}_{\boldsymbol{\theta}}(\mathbf{z}_t,t)-\boldsymbol{\epsilon}_t
\right\|_2^2
\right].
\end{equation}
This objective is closely related to score matching, since predicting $\boldsymbol{\epsilon}_t$ is equivalent to estimating the score $\nabla_{\mathbf{z}_t}\log p(\mathbf{z}_t)$. In the variance-preserving formulation \cite{sde}, the reverse dynamics can be written as
\begin{equation}
	\mathbf{z}_{t-1}
	=
	\left(1+\frac{1}{2}\beta_t\right)\mathbf{z}_t
	+
	\beta_t \nabla_{\mathbf{z}_t}\log p(\mathbf{z}_t)
	+
	\sqrt{\beta_t}\,\boldsymbol{\epsilon},
\end{equation}
where $\boldsymbol{\epsilon}\sim\mathcal{N}(0,\mathbf{I})$.

For conditional generation, the objective is to sample from the conditional distribution associated with a target condition $\mathbf{y}$. Accordingly, the reverse step becomes
\begin{equation}
\label{eq:cond_sde}
	\mathbf{z}_{t-1}
	=
	\left(1+\frac{1}{2}\beta_t\right)\mathbf{z}_t
	+
	\beta_t \nabla_{\mathbf{z}_t}\log p(\mathbf{z}_t|\mathbf{y})
	+
	\sqrt{\beta_t}\,\boldsymbol{\epsilon}.
\end{equation}
Using Bayes' rule, the conditional score can be decomposed into an unconditional prior term and a condition-dependent correction term:
\begin{equation}
\label{eq:correction_grad}
\begin{aligned}
\nabla_{\mathbf{z}_t}\log p(\mathbf{z}_t \mid \mathbf{y})
&=
\nabla_{\mathbf{z}_t}\log p(\mathbf{z}_t)
+
\nabla_{\mathbf{z}_t}\log p(\mathbf{y} \mid \mathbf{z}_t) \\
&=
\nabla_{\mathbf{z}_t}\log p(\mathbf{z}_t)
-
\lambda \nabla_{\mathbf{z}_t}\mathcal{L}_t(\mathbf{z}_t,\mathbf{y}),
\end{aligned}
\end{equation}
where $\lambda$ controls the contribution of the guidance term. Substituting \cref{eq:correction_grad} into \cref{eq:cond_sde} yields
\begin{equation}
\label{eq:cond_diffusion}
\mathbf{z}_{t-1}
=
\mathbf{m}_t
-
\eta_t \nabla_{\mathbf{z}_t}\mathcal{L}_t(\mathbf{z}_t,\mathbf{y}),
\end{equation}
where
\begin{equation}
\label{eq:mt}
\mathbf{m}_t
=
\left(1+\frac{1}{2}\beta_t\right)\mathbf{z}_t
+
\beta_t \nabla_{\mathbf{z}_t}\log p(\mathbf{z}_t)
+
\sqrt{\beta_t}\,\boldsymbol{\epsilon},
\end{equation}
and $\eta_t$ denotes the effective guidance scale at timestep $t$. In practice, the gradient term in \cref{eq:cond_diffusion} is computed by backpropagating the loss through the guidance model together with the diffusion network, which makes the framework compatible with a broad range of task-specific objectives.

\subsection{Ambiguity-Aware Knowledge Graph (AAKG)}

\noindent \textbf{Sense disambiguation.} Natural language prompts often contain polysemous words whose meanings vary with context, leading to potential semantic ambiguity during generation. To resolve this issue, for each ambiguous token $w$, we first retrieve a set of candidate senses:
\begin{equation}
w \rightarrow \{s_1, s_2, \cdots, s_K\},
\end{equation}
where $\{s_i\}_{i=1}^{K}$ denotes the candidate meanings associated with $w$. We then identify the most contextually consistent sense by evaluating each candidate under the prompt condition $\mathbf{y}$ and the external knowledge source $\mathcal{K}$:
\begin{equation}
s^* = \arg\max_{s_i} \text{Score}(s_i \mid \mathbf{y}, \mathcal{K}).
\end{equation}
In this way, the selected sense $s^*$ provides a semantically grounded interpretation for ambiguous words before subsequent structural modeling.

\noindent \textbf{Graph construction.} Based on the disambiguated prompt semantics, we construct an ambiguity-aware semantic graph
\begin{equation}
\mathcal{G} = (\mathcal{V}, \mathcal{E}),
\end{equation}
where the node set $\mathcal{V}$ consists of semantic elements, including entities and their associated attributes, and the edge set $\mathcal{E}$ characterizes the relationships among them. In addition to positive semantic associations, $\mathcal{E}$ also includes negative constraints to explicitly suppress incompatible or misleading bindings. Such a graph representation enables the prompt semantics to be organized in a structured form, which is more suitable for downstream guidance than using raw text alone.

\noindent \textbf{Semantic grouping.} To facilitate targeted guidance at different denoising stages, we further decompose $\mathcal{G}$ into four semantic groups:
\begin{equation}
\mathcal{T} = \{T_{\mathrm{id}}, T_{\mathrm{rel}}, T_{\mathrm{attr}}, T_{\mathrm{neg}}\}.
\end{equation}
Specifically, $T_{\mathrm{id}}$ denotes identity-related semantics, $T_{\mathrm{rel}}$ captures relational semantics between entities, $T_{\mathrm{attr}}$ contains attribute-level descriptions, and $T_{\mathrm{neg}}$ represents negative semantic constraints. This decomposition allows different types of semantic information to be modeled and injected separately, improving the controllability and precision of the guidance process.

\subsection{Structured Semantic Injection Module (SSIM)}
Although the ambiguity-aware knowledge graph provides structured semantic representations, 
a key challenge is how to effectively incorporate such structured signals into the diffusion process. 
Directly conditioning on global embeddings often leads to semantic competition and entanglement, 
making it difficult to enforce fine-grained control over identity, relations, and attributes. 
To address this issue, we propose a unified framework that integrates structured semantic injection 
with explicit semantic control objectives.

\vspace{0.1cm}



We first encode each semantic group derived from the knowledge graph into disentangled embeddings:
\begin{equation}
T_k = E_k(\mathcal{G}), \quad k \in \{\mathrm{id, rel, attr, neg}\},
\end{equation}
where $E_k(\cdot)$ denotes the semantic encoder for identity, relational, attribute, and negative constraints, respectively.

Instead of applying semantic guidance only at the output level, 
we inject structured semantic signals into intermediate layers of the diffusion backbone. 
Specifically, for the latent feature at timestep $t$ and layer $l$, we perform:
\begin{equation}
\mathbf{z}_t^{(l)} = \mathbf{z}_t^{(l)} + \sum_k \lambda_k(t) T_k,
\end{equation}
where $\lambda_k(t)$ controls the strength of each semantic component.

\vspace{0.2cm}

To further enforce semantic consistency, we formulate a multi-objective optimization problem 
based on cross-attention maps, which reflect the alignment between text tokens and image regions.

We denote $M_u$ as the cross-attention map corresponding to token $u$, 
and $f(\cdot,\cdot)$ as the cosine similarity function. 
$\mathbf{S}_{id}, \mathbf{S}_{rel}, \mathbf{S}_{attr}, \mathbf{S}_{neg}$ 
denote identity, relation, attribute, and negative semantic pair sets, respectively.

\vspace{0.1cm}

\noindent \textbf{Identity consistency.}
\begin{equation}
\mathcal{L}_{id} = - \frac{1}{|\mathbf{S}_{id}|} 
\sum_{(u,v) \in \mathbf{S}_{id}} f(M_u, M_v).
\end{equation}

\noindent \textbf{Relational consistency.}
\begin{equation}
\mathcal{L}_{rel} = - \frac{1}{|\mathbf{S}_{rel}|}
\sum_{(u,v) \in \mathbf{S}_{rel}} f(M_u, M_v).
\end{equation}

\noindent \textbf{Attribute alignment.}
\begin{equation}
\begin{aligned}
\mathcal{L}_{attr} = & -\frac{1}{|\mathbf{S}_{attr}^{pos}|}
\sum_{(u,v) \in \mathbf{S}_{attr}^{pos}} f(M_u, M_v) \\
& + \frac{1}{|\mathbf{S}_{attr}^{neg}|}
\sum_{(u,v) \in \mathbf{S}_{attr}^{neg}} f(M_u, M_v).
\end{aligned}
\end{equation}

\noindent \textbf{Negative constraint.}
\begin{equation}
\mathcal{L}_{neg} = \frac{1}{|\mathbf{S}_{neg}|}
\sum_{(u,v) \in \mathbf{S}_{neg}} f(M_u, M_v).
\end{equation}

\noindent \textbf{Temporal consistency.}
\begin{equation}
\mathcal{L}_{temp} = \frac{1}{|\mathcal{S}|}
\sum_{(s,l)\in \mathcal{S}} 
\left( f(M_s^t, M_l^t) - f(M_s^{t-1}, M_l^{t-1}) \right)^2.
\end{equation}

\vspace{0.2cm}

We combine all semantic objectives into a unified formulation:
\begin{equation}
\begin{aligned}
\mathcal{L}_{total}(t) = &
\lambda_{id}(t)\mathcal{L}_{id} +
\lambda_{rel}(t)\mathcal{L}_{rel} \\
& + \lambda_{attr}(t)\mathcal{L}_{attr}
- \lambda_{neg}(t)\mathcal{L}_{neg} \\
& + \lambda_{temp}(t)\mathcal{L}_{temp}.
\end{aligned}
\end{equation}

\noindent The latent variable is updated via gradient-based guidance:
\begin{equation}
\mathbf{z}_{t-1} = \mathbf{m}_t - \eta_t \nabla_{\mathbf{z}_t} \mathcal{L}_{total}(t).
\end{equation}

\vspace{0.1cm}

\subsection{Temporal-Aware Semantic Control (TASC)}
\label{subsec:tasc}

Different semantic factors play distinct roles across the denoising trajectory. 
In the early stage, the latent remains highly noisy, and the generation process is primarily responsible for establishing the global semantic identity and suppressing incorrect concepts. 
In the middle stage, the coarse structure becomes clearer, making this phase more suitable for modeling inter-entity relations. 
In the late stage, the image structure is largely stabilized, and the denoising process becomes more sensitive to fine-grained attributes and local details. 
Therefore, applying all semantic constraints uniformly across timesteps is suboptimal, as it ignores the stage-dependent dynamics of diffusion. 
To address this issue, we introduce a Temporal-Aware Semantic Control (TASC) module that dynamically schedules the contribution of different semantic objectives over time.

Specifically, we assign timestep-dependent weights to identity, relational, attribute, and negative semantic constraints:
\begin{equation}
\begin{aligned}
\lambda_{\mathrm{id}}(t)   &= \exp(-\alpha t), \\
\lambda_{\mathrm{rel}}(t)  &= \exp\left(-\frac{(t-\mu)^2}{2\sigma^2}\right), \\
\lambda_{\mathrm{attr}}(t) &= 1 - \exp(-\beta t), \\
\lambda_{\mathrm{neg}}(t)  &= \gamma \exp(-\delta t).
\end{aligned}
\end{equation}
where $\lambda_{\mathrm{id}}(t)$ emphasizes semantic identity in early denoising steps, 
$\lambda_{\mathrm{rel}}(t)$ reaches its maximum in the middle stage to guide structural and relational composition, 
$\lambda_{\mathrm{attr}}(t)$ gradually increases to enhance fine-grained attributes in later stages, 
and $\lambda_{\mathrm{neg}}(t)$ imposes strong early suppression on incorrect semantic interpretations. 
In this way, TASC aligns semantic guidance with the intrinsic phase-dependent behavior of diffusion.

\vspace{0.1cm}

\begin{figure*}[p]
    \centering
    \includegraphics[width=\textwidth,height=0.82\textheight,keepaspectratio]{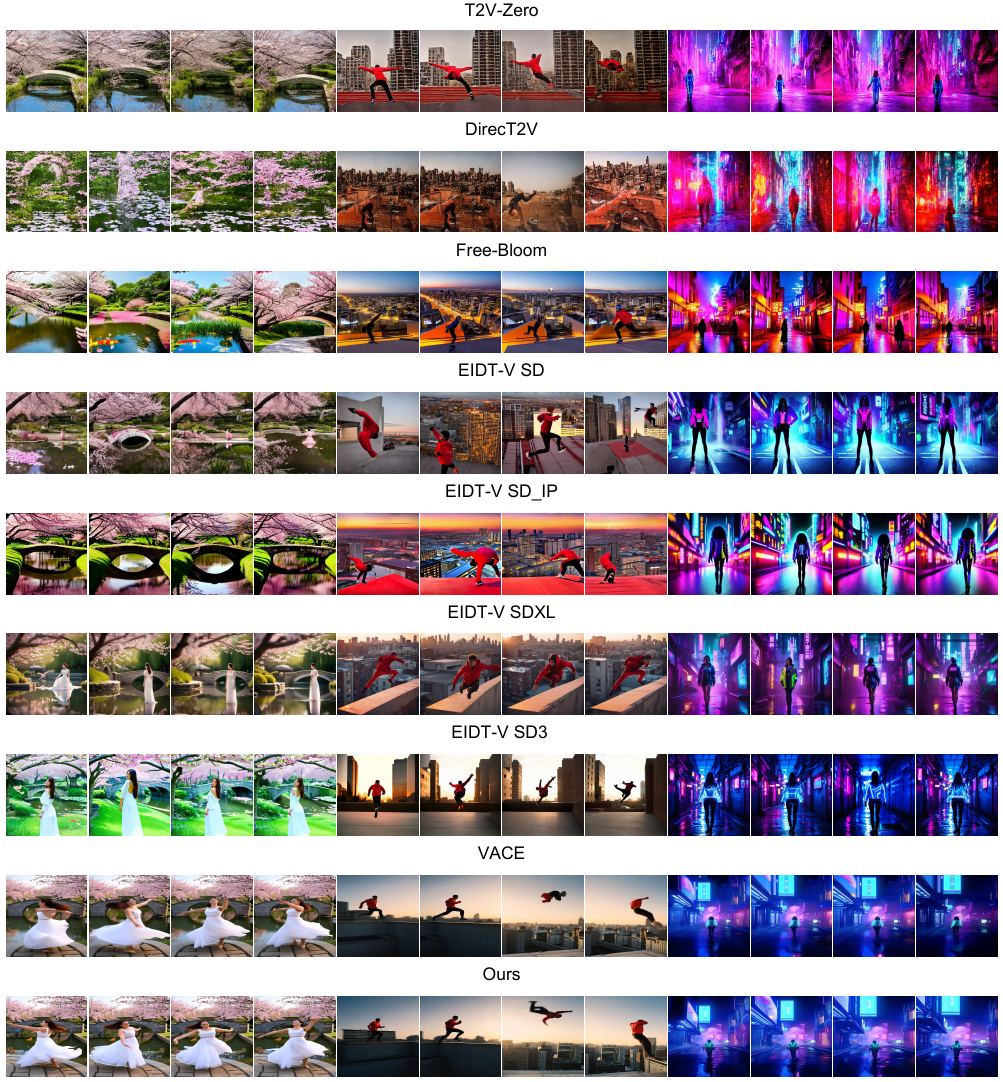}
    \caption{\textbf{Qualitative comparison with other methods.}                From left to right, the prompts are: \textbf{(I)} ``A woman in a white dress performs a dynamic, sweeping dance, spinning rapidly with her arms outstretched and her dress billowing outward, in a tranquil garden with peach blossoms in full bloom. A stone arch bridge and still pond behind her, pink petals drifting in the breeze. Koi swim beneath reflections of blue sky and blossoms. Soft warm spring light. Static camera, medium shot with shallow depth of field''; \textbf{(II)} ``Cinematic shot, a parkour athlete in red sprints across rooftops, leaps over a wide gap between buildings with body fully stretched, lands into a roll and backflips off a wall. Jacket flares in the wind. Camera tracks fast from the side, dusk skyline glowing orange-gold. Dynamic wide to medium shots, intense speed and impact''; \textbf{(III)} ``Cyberpunk style, a woman in a glowing tech jacket walks through a neon-lit alley at night in a futuristic city. Holographic billboards cast blue-purple light, wet streets reflect neon colors, steam rises from below. Deep blue, electric purple, and neon pink palette. Camera slowly pushes forward, medium shot, cinematic sci-fi atmosphere.''}
    \label{fig:comparison}
\end{figure*}

\section{Experiments}
\subsection{Implementation Details}
We build KGEdit on VACE~\cite{vace}, using Wan 2.1 1.3B as the video diffusion backbone. 
The backbone is a 30-layer DiT with a hidden size of 1536, 12 attention heads, and an FFN dimension of 8960, running in bfloat16 precision. 
UMT5-XXL is used as the text encoder with a maximum length of 512 tokens. 
Unless otherwise specified, all videos are generated at $480\times832$ resolution with 81 frames at 16 FPS, using the UniPC sampler with 50 denoising steps and a classifier-free guidance scale of 5.0. 
No additional training or fine-tuning is performed; all modules operate in a training-free manner at inference time.

For AAKG, we use Qwen2.5-3B-Instruct for sense disambiguation and structured triplet extraction. 
The generated subject-predicate-object triplets are parsed and categorized into four semantic groups, i.e., $T_{\mathrm{id}}$, $T_{\mathrm{rel}}$, $T_{\mathrm{attr}}$, and $T_{\mathrm{neg}}$, based on predefined predicate vocabularies. 
For SSIM, structured semantic signals are injected into 15 even-indexed DiT layers, i.e., layers $\{0,2,\ldots,28\}$, following the default VACE context injection positions. 
For TASC, we set $\alpha{=}4.0$, $\mu{=}0.5$, $\sigma{=}0.18$, $\beta{=}4.0$, $\gamma{=}1.2$, and $\delta{=}5.0$. 
All experiments are conducted on a single NVIDIA A100 GPU.

\noindent \textbf{Datasets.}
We evaluate KGEdit on VBench~\cite{huang2023vbench}, which provides standardized prompts and comprehensive automatic metrics for video generation. All compared methods use the same prompt set for fair comparison.

\noindent \textbf{Baselines.} We compare our method with eight representative baselines: 
\textbf{1) T2V-Zero}~\cite{khachatryan2023t2video}: a classical zero-shot text-to-video generation method built upon pretrained text-to-image diffusion models; 
\textbf{2) DirecT2V}~\cite{hong2023large}: a training-free framework that improves video generation via frame-wise prompt decomposition; 
\textbf{3) Free-Bloom}~\cite{huang2023free}: a zero-shot and training-free video generation method with enhanced temporal consistency; 
\textbf{4) EIDT-V SD}~\cite{jagpal2025eidt}: the Stable Diffusion 1.5-based version of EIDT-V for model-agnostic, zero-shot, and training-free video generation; 
\textbf{5) EIDT-V SD3}~\cite{jagpal2025eidt}: the SD3-based variant of EIDT-V with a stronger generative backbone; 
\textbf{6) EIDT-V SDXL}~\cite{jagpal2025eidt}: the SDXL-based variant of EIDT-V for improved scalability and generation quality; 
\textbf{7) EIDT-V SD\_IP}~\cite{jagpal2025eidt}: an EIDT-V variant combined with IP-Adapter to incorporate image-prompt guidance; and 
\textbf{8) VACE}~\cite{vace}: a unified video generation and editing framework supporting multiple conditional video creation tasks.

\noindent \textbf{Evaluation.}
We adopt VBench~\cite{huang2023vbench} as the primary automatic evaluation protocol, which assesses generated videos across multiple complementary dimensions.
Specifically, we report seven metrics spanning two categories: (i) \emph{video quality}, including Aesthetic Quality and Imaging Quality, which measure visual appearance and perceptual fidelity; and (ii) \emph{video consistency}, including Background Consistency, Subject Consistency, Motion Smoothness, Temporal Flickering, and Overall Consistency, which capture temporal coherence and semantic alignment.
We further conduct a randomized blind evaluation (user study) in which 18 raters score videos generated by different methods on representative prompts using a 5-point scale, and the mean score is mapped to a 100-point scale. covering prompt following, temporal consistency, and overall video quality.

\subsection{Main Results}
\textbf{Quantitative Results.}
Table~\ref{tab:vbench_comparison} reports the VBench results. Our method achieves the best average score of \textbf{79.62} across all seven metrics, surpassing all baselines, including the unified VACE framework.

For temporal consistency, our method performs best on four metrics: Background Consistency (98.89), Motion Smoothness (99.19), Subject Consistency (99.20), and Temporal Flickering (98.61). These results verify the effectiveness of TASC in stabilizing the temporal propagation of key attributes, producing more coherent and flicker-free videos.

Compared with VACE, our method improves Overall Consistency from 23.46 to 26.83 (+3.37), highlighting AAKG's value in resolving ambiguity and providing structured guidance. It also improves Aesthetic Quality (+0.65) and Imaging Quality (+1.54), suggesting that SSIM enhances visual fidelity through finer-grained control.

Among the training-free baselines, EIDT-V SD3 achieves the best Aesthetic Quality (68.61) due to its stronger SD3 backbone, while EIDT-V SDXL attains the highest Overall Consistency (28.99). However, both methods perform notably worse on temporal consistency metrics, with Subject Consistency below 90 and Temporal Flickering below 93, indicating a clear trade-off between frame-level quality and temporal stability. By contrast, our method preserves strong temporal coherence while maintaining competitive overall quality, achieving a better balance between video quality and temporal stability. Free-Bloom obtains the best Imaging Quality (74.16) but shows weaker cross-frame consistency, further highlighting the difficulty of optimizing both aspects jointly.

\begin{table*}[t]
\centering
\cvprtablecaptionfull{\textbf{Quantitative evaluations on VBench.} We report seven automatic VBench metrics together with three human-study scores. Average is computed over the seven automatic VBench metrics. Bold indicates the best result.}
\label{tab:vbench_comparison}
\vspace{0.4em}

\scriptsize
\renewcommand{\arraystretch}{1.05}

\begin{tabular*}{\textwidth}{@{\extracolsep{\fill}}lccccccccccc@{}}
\toprule
\multirow{2}{*}{\textbf{Model}}
& \multicolumn{8}{c}{\textbf{Video Quality \& Video Consistency}}
& \multicolumn{3}{c}{\textbf{User Study}} \\
\cmidrule(lr){2-9} \cmidrule(lr){10-12}
& \rotheader{Aesthetic\\Quality}
& \rotheader{Background\\Consistency}
& \rotheader{Imaging\\Quality}
& \rotheader{Motion\\Smoothness}
& \rotheader{Overall\\Consistency}
& \rotheader{Subject\\Consistency}
& \rotheader{Temporal\\Flickering}
& \rotheader{Average}
& \rotheader{Prompt\\Following}
& \rotheader{Temporal\\Consistency}
& \rotheader{Video\\Quality} \\
\midrule
T2V-Zero~\cite{khachatryan2023t2video}      & 65.67 & 96.07 & 71.99 & 88.08 & 25.46 & 94.00 & 85.31 & 75.23 & 67.80 & 66.20 & 68.40 \\
DirecT2V~\cite{hong2023large}      & 56.23 & 91.77 & 71.24 & 86.48 & 24.37 & 81.46 & 85.18 & 70.96 & 58.30 & 55.40 & 57.90 \\
Free-Bloom~\cite{huang2023free}    & 65.07 & 93.79 & \textbf{74.16} & 90.78 & 26.28 & 85.89 & 89.67 & 75.09 & 64.50 & 63.80 & 66.30 \\
EIDT-V SD~\cite{jagpal2025eidt}     & 62.23 & 85.25 & 71.67 & 91.11 & 25.90 & 67.92 & 90.14 & 70.60 & 60.70 & 57.60 & 59.50 \\
EIDT-V SD3~\cite{jagpal2025eidt}    & \textbf{68.61} & 95.10 & 72.91 & 93.43 & 27.70 & 89.89 & 92.63 & 77.18 & 71.20 & 68.90 & 72.80 \\
EIDT-V SDXL~\cite{jagpal2025eidt}   & 67.78 & 94.76 & 71.85 & 91.46 & \textbf{28.99} & 86.92 & 90.05 & 75.97 & 72.60 & 65.70 & 70.10 \\
EIDT-V SD\_IP~\cite{jagpal2025eidt} & 63.75 & 93.77 & 68.80 & 91.37 & 25.16 & 86.67 & 90.23 & 74.25 & 63.10 & 61.50 & 64.70 \\
VACE~\cite{vace}          & 64.31 & 98.84 & 68.09 & 99.08 & 23.46 & 99.05 & 98.56 & 78.77 & 79.40 & 83.10 & 78.30 \\
\textbf{Ours} & 64.96 & \textbf{98.89} & 69.63 & \textbf{99.19} & 26.83 & \textbf{99.20} & \textbf{98.61} & \textbf{79.62} & \textbf{83.50} & \textbf{87.20} & \textbf{82.60} \\
\bottomrule
\end{tabular*}
\vspace{-1.6em}
\end{table*}

\textbf{Qualitative Results.}
Fig.~\ref{fig:comparison} presents qualitative comparisons with representative baselines across three diverse prompts involving complex motion, multiple attributes, and stylized scenes.
As shown in Fig.~\ref{fig:comparison}, existing training-free methods such as T2V-Zero and Free-Bloom tend to generate videos with noticeable temporal flickering and inconsistent subject appearances, particularly under prompts involving rapid motion (prompt II) or fine-grained environmental descriptions (prompt III).
The EIDT-V variants improve per-frame visual quality but still exhibit cross-frame semantic drift, where attributes such as clothing color or background elements shift across the sequence.
The base VACE model achieves stable temporal consistency but occasionally produces results that deviate from the intended semantics, especially when the prompt contains ambiguous or multi-attribute descriptions.

In contrast, our method generates temporally coherent videos with more precise semantic alignment.
For instance, in prompt (I) involving a dancing woman with specific dress and environmental descriptions, our method correctly preserves all specified attributes---dress appearance, garden elements, and lighting conditions---throughout the sequence.
In prompt (III) with a cyberpunk scene, our approach accurately renders the specified color palette (deep blue, electric purple, neon pink) and atmospheric details (holographic billboards, wet reflections, rising steam), whereas baselines tend to lose fine-grained attributes or produce inconsistent lighting across frames.
These observations confirm that the structured semantic guidance provided by the AAKG, combined with the targeted injection of SSIM and the temporal scheduling of TASC, enables more precise and temporally stable video generation under complex prompts.

\vspace{-0.8em}
\subsection{Ablation Studies}
We conduct ablations to analyze the contributions of AAKG, each triplet group, and TASC scheduling.

\noindent \textbf{A. Effectiveness of Ambiguity-Aware Knowledge Graph.}
We first evaluate the overall effect of the proposed AAKG module by comparing our full model (with AAKG) against a variant without AAKG (w/o AAKG) under a multi-round interactive editing setting. Two challenging categories are examined: (i) ambiguous prompts containing polysemous words, and (ii) complex prompts requiring precise multi-attribute binding. In each case, we show that AAKG enables the model to produce correct results within a single editing round, whereas the variant without AAKG fails even after multiple rounds of prompt refinement.

\begin{figure}[t]
    \centering
    \includegraphics[width=\columnwidth]{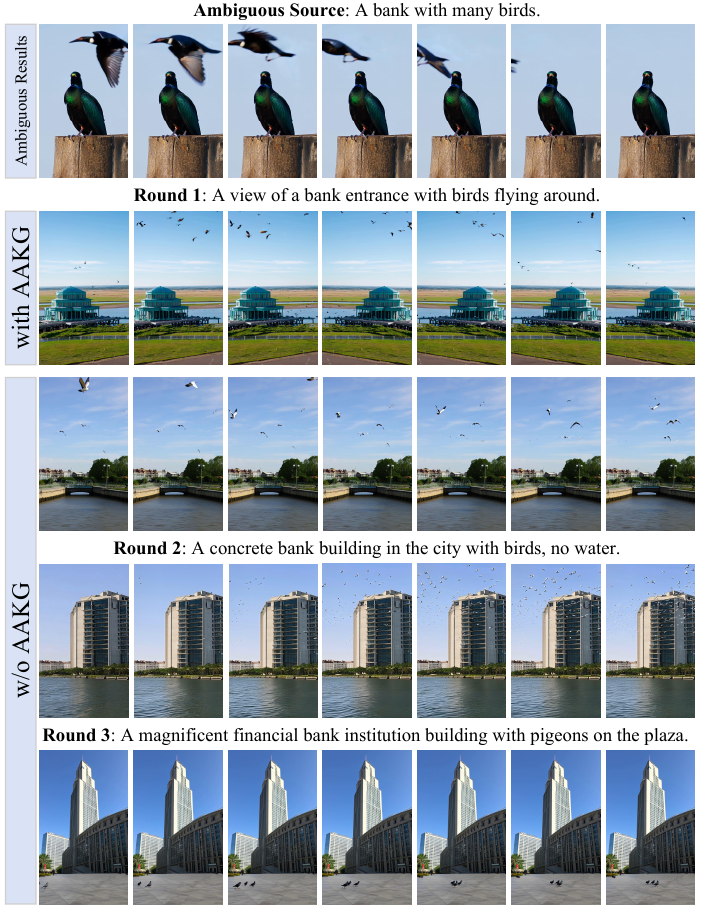}
    \vspace{-1.2em}
    \caption{\textbf{Ablation on AAKG guidance (ambiguity).} The prompt ``A bank with many birds'' contains the polysemous word ``bank.'' Without AAKG guidance, the model persistently misinterprets ``bank'' as a financial institution even after multiple rounds of prompt refinement. Our method, aided by structured disambiguation ($T_{\mathrm{id}}$: bank = riverbank; $T_{\mathrm{neg}}$: NOT financial\_institution), correctly generates a natural riverbank scene with birds.}
    \label{fig:ab_kg1}
    \vspace{-2.0em}
\end{figure}

\begin{figure}[t]
    \centering
    \includegraphics[width=\columnwidth]{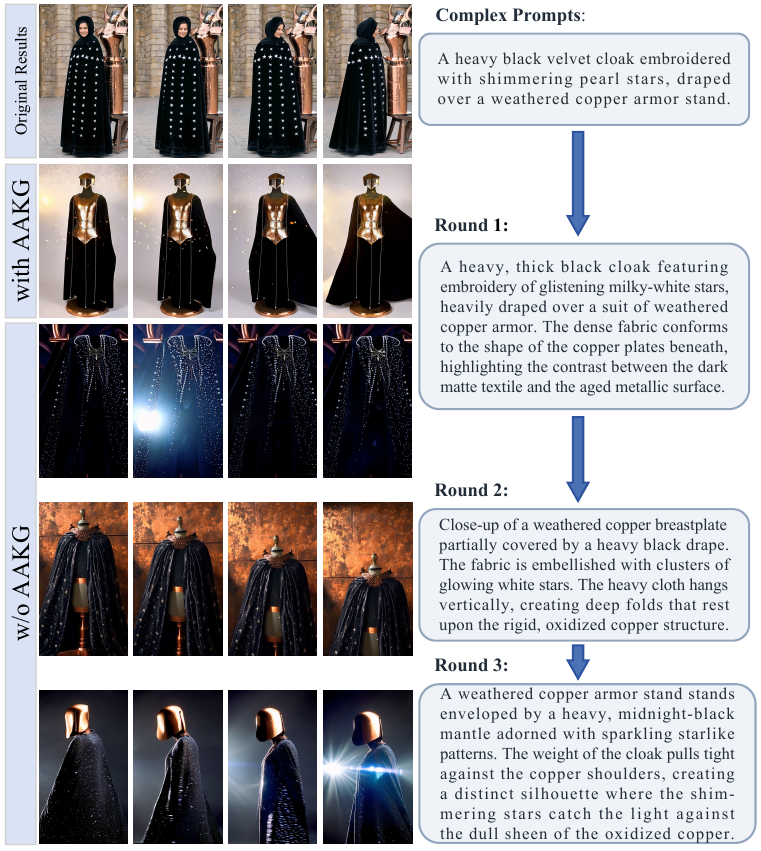}
    \vspace{-1.2em}
    \caption{\textbf{Ablation on AAKG guidance (attribute binding).} Under a complex multi-attribute prompt describing ``a heavy black velvet cloak embroidered with shimmering pearl stars, draped over a weathered copper armor stand,'' the baseline without AAKG produces attribute leakage and incorrect material binding across multiple refinement rounds, while our method preserves correct attribute bindings through structured semantic decomposition.}
    \label{fig:ab_kg2}
    \vspace{-0.4em}
\end{figure}

As shown in Fig.~\ref{fig:ab_kg1}, the prompt ``A bank with many birds'' contains the polysemous word ``bank,'' which can refer to either a riverbank or a financial institution. Without AAKG guidance, the baseline suffers from persistent semantic drift: even after multiple rounds of interactive prompt refinement, it continues to generate financial bank buildings instead of a natural riverbank scene. In contrast, our method explicitly disambiguates the concept through the AAKG, producing the identity triplet $T_{\mathrm{id}}$: (\textit{bank}, \textit{is}, \textit{riverbank}) and the negative constraint $T_{\mathrm{neg}}$: (\textit{bank}, \textit{not\_instance\_of}, \textit{financial\_institution}), thereby correctly guiding the generation toward a riverbank scene with birds in a single round.

Fig.~\ref{fig:ab_kg2} further demonstrates the advantage of our method in complex multi-attribute scenarios. The prompt describes ``a heavy black velvet cloak embroidered with shimmering pearl stars, draped over a weathered copper armor stand,'' involving multiple fine-grained attributes such as material (velvet, copper), color (black), embellishment (pearl stars), and spatial relation (draped over). Without AAKG, the baseline produces attribute leakage and incorrect material binding even after multiple rounds of prompt refinement. Our method decomposes the scene into structured attribute triplets such as (\textit{cloak}, \textit{has\_material}, \textit{black\_velvet}) and (\textit{cloak}, \textit{has\_embroidery}, \textit{shimmering\_pearl\_stars}), along with relational triplets such as (\textit{cloak}, \textit{draped\_over}, \textit{armor\_stand}), enabling disentangled and faithful generation.

\vspace{0.2cm}
\noindent \textbf{B. Ablation on Triplet Groups.}
To validate the contribution of each semantic group, we systematically remove individual triplet groups and evaluate the resulting performance. 
Table~\ref{tab:ablation_triplet} summarizes the quantitative results.

\begin{table}[h]
  \centering
  \cvprtablecaptionhalf{\textbf{Ablation study on Triplet Groups and TASC.} We report Semantic Accuracy (SA), Attribute Binding accuracy (AB), Temporal Consistency (TC), and FVD. $\checkmark$ indicates the component is included; best results are \textbf{bolded}.}
  \label{tab:ablation_triplet}
  \vspace{0.3em}
  \resizebox{\linewidth}{!}{%
  \begin{tabular}{l|cccc|cccc}
    \toprule
    \multirow{2}{*}{Configuration} & \multicolumn{4}{c|}{Components} & \multicolumn{4}{c}{Metrics} \\
    & $T_{id}$ & $T_{rel}$ & $T_{attr}$ & $T_{neg}$ & SA$\uparrow$ & AB$\uparrow$ & TC$\uparrow$ & FVD$\downarrow$ \\
    \midrule
    Baseline (no KG) &  &  &  &  & 62.3 & 58.7 & 71.2 & 312.4 \\ \midrule
    w/o $T_{id}$  &  & \checkmark & \checkmark & \checkmark & 71.6 & 72.5 & 76.8 & 258.1 \\
    w/o $T_{rel}$  & \checkmark &  & \checkmark & \checkmark & 78.5 & 71.3 & 73.4 & 245.6 \\
    w/o $T_{attr}$ & \checkmark & \checkmark &  & \checkmark & 79.2 & 63.8 & 78.1 & 241.3 \\
    w/o $T_{neg}$  & \checkmark & \checkmark & \checkmark &  & 74.1 & 74.2 & 77.5 & 271.8 \\
    w/o TASC       & \checkmark & \checkmark & \checkmark & \checkmark & 80.3 & 76.1 & 74.9 & 234.5 \\ \midrule
    Full (Ours) & \checkmark & \checkmark & \checkmark & \checkmark & \textbf{85.7} & \textbf{82.4} & \textbf{83.6} & \textbf{198.2} \\
    \bottomrule
  \end{tabular}}
  \vspace{-1.8em}
\end{table}

Several observations can be drawn from the results.
(1)~\emph{Each triplet group contributes uniquely and irreplaceably.} Removing any single group leads to noticeable performance degradation, confirming that identity, relation, attribute, and negative constraints address distinct semantic aspects of the generation process.
(2)~\emph{$T_{id}$ is the most critical for semantic accuracy.} The w/o~$T_{id}$ variant exhibits the largest SA drop ($-$14.1 compared to Full), indicating that identity disambiguation is the foundational step: without correctly grounding the core concept (e.g., ``bank = riverbank''), downstream semantics inevitably drift.
(3)~\emph{$T_{neg}$ is essential for ambiguity resolution.} Removing negative constraints causes the second-largest SA decline ($-$11.6), validating that explicitly suppressing incorrect interpretations (e.g., ``NOT financial\_institution'') is critical for polysemous inputs.
(4)~\emph{$T_{attr}$ dominates attribute binding fidelity.} The w/o~$T_{attr}$ configuration shows the steepest AB decline ($-$18.6 from Full), demonstrating that structured attribute triplets such as (\textit{has\_material}, \textit{black\_velvet}) are essential for preventing attribute leakage across entities.
(5)~\emph{$T_{rel}$ maintains spatial and compositional coherence.} Removing relational triplets degrades TC by $-$10.2, confirming that explicit spatial constraints such as (\textit{draped\_over}, \textit{armor\_stand}) stabilize inter-entity composition across frames.

\vspace{0.2cm}
\noindent \textbf{C. Effectiveness of TASC.}
We further analyze the contribution of the Temporal-Aware Semantic Control module by comparing our full method with a variant using uniform temporal weights (w/o~TASC row in Table~\ref{tab:ablation_triplet}).

\begin{figure}[t]
    \centering
    \includegraphics[width=\columnwidth]{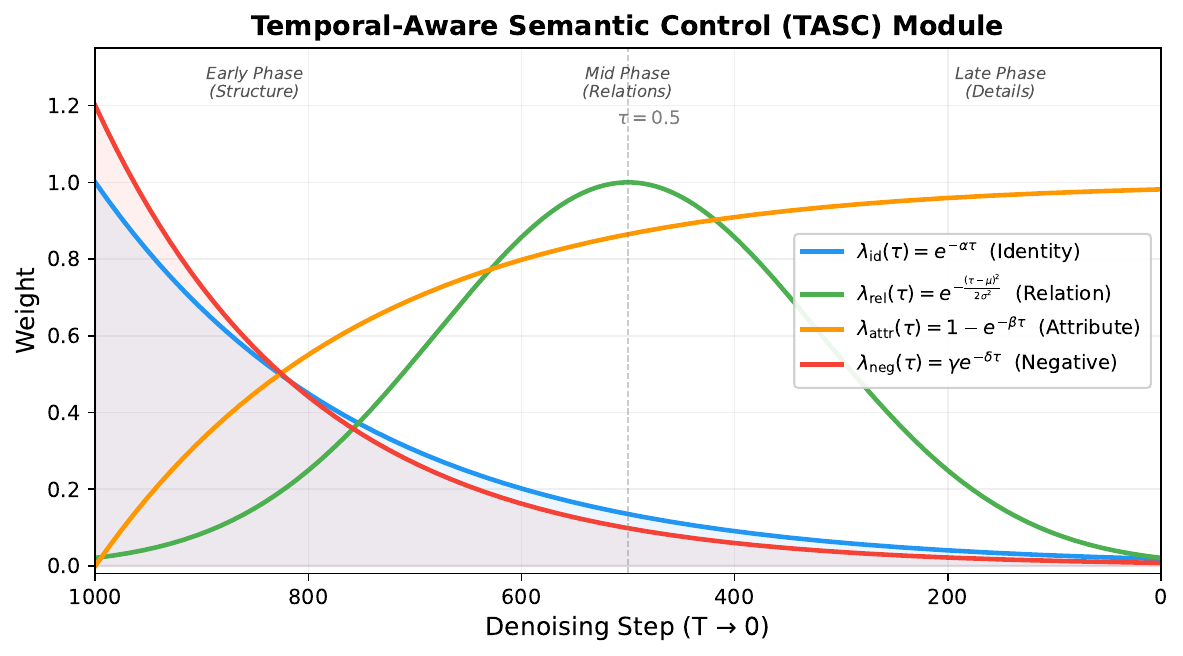}
    \vspace{-1.8em}
    \caption{\textbf{Temporal weight curves of the TASC module.} $\lambda_{\mathrm{id}}$ and $\lambda_{\mathrm{neg}}$ dominate early denoising steps for structural grounding and disambiguation, $\lambda_{\mathrm{rel}}$ peaks at mid-stage for compositional reasoning, and $\lambda_{\mathrm{attr}}$ grows toward later steps for fine-grained detail refinement.}
    \label{fig:tasc_weights}
    \vspace{-1.4em}
\end{figure}

Fig.~\ref{fig:tasc_weights} visualizes the four temporal weight functions.
The design rationale is motivated by the stage-dependent behavior of diffusion models: early denoising steps determine global structure and semantic identity, middle steps establish inter-entity spatial composition, and late steps refine fine-grained textures and attributes.

As shown in Table~\ref{tab:ablation_triplet}, the w/o~TASC variant, which applies all four triplet groups with uniform weights throughout denoising, underperforms the full method by $-$5.4 in SA, $-$6.3 in AB, and $-$8.7 in TC. This confirms that different semantic components require distinct temporal emphasis, while uniform weighting fails to align structured guidance with the dynamics of denoising. Notably, the TC drop ($-$8.7) shows that temporal scheduling is crucial for consistent attribute propagation: without TASC, identity constraints useful in early steps may interfere with late-stage detail refinement, causing flickering and attribute instability.

Overall, these ablations demonstrate that KGEdit benefits from (i) structured decomposition into four complementary triplet groups, (ii) explicit negative constraints for disambiguation, and (iii) dynamic temporal scheduling aligned with the diffusion process.

\vspace{-0.6em}
\section{Conclusion}
\noindent KGEdit is a training-free framework for structured semantic control in video generation and editing. It uses an ambiguity-aware knowledge graph to disambiguate prompts into identity, relation, attribute, and negative constraints, injects them into key diffusion-Transformer layers via SSIM, and schedules their effects with TASC across denoising stages. Experiments and ablations show improved semantic alignment, attribute binding, and temporal consistency over representative training-free methods and unified frameworks. 

\noindent\textbf{Limitations.} Current limitations include LLM-based graph construction, fixed injection layers, and hand-designed schedules; future work will explore adaptive graphs, learned scheduling, and lightweight guidance.


\bibliographystyle{IEEEtran}
\bibliography{ref}

\vfill

\end{document}